%% file: main.tex
\documentclass[lettersize,journal]{IEEEtran}
\usepackage{amsmath,amsfonts}
\pdfoutput=1
\usepackage{array}
\usepackage[caption=false,font=normalsize,labelfont=sf,textfont=sf]{subfig}
\usepackage{textcomp}
\usepackage{stfloats}
\usepackage{url}
\usepackage{verbatim}
\usepackage{graphicx}
\usepackage{booktabs}
\usepackage{cite}
\usepackage{color,xcolor}
\usepackage{dsfont}
\usepackage[ruled,linesnumbered]{algorithm2e}
\usepackage{xurl}

\usepackage{times}
\usepackage{epsfig}
\usepackage{graphicx}
\usepackage{amsmath}
\usepackage{amssymb}
\usepackage{booktabs}
\usepackage[ruled,linesnumbered]{algorithm2e}
\usepackage{silence}
\usepackage{threeparttable}
\usepackage{dsfont}
\usepackage{multirow}
\DeclareMathOperator*{\argmax}{argmax}
\DeclareMathOperator*{\argmin}{argmin}

\begin{document}

\title{Stealthy Physical Masked Face Recognition Attack via Adversarial Style Optimization}

\author{Huihui~Gong,~Minjing Dong,~Siqi~Ma,~Seyit~Camtepe,~\IEEEmembership{Senior~Member,~IEEE,}~Surya~Nepal,~and ~Chang~Xu,~\IEEEmembership{Member,~IEEE}
\IEEEcompsocitemizethanks{\IEEEcompsocthanksitem H. Gong, M. Dong and C. Xu are with the School of Computer Science, Faculty of Engineering, The University of Sydney, Sydney, NSW 2008, Australia (e-mail: hgon9611@uni.sydney.edu.au; mdon0736@uni.sydney.edu.au; c.xu@sydney.edu.au).\protect
\IEEEcompsocthanksitem S. Ma is with the School of Engineering and Information Technology, University of New South Wales Canberra, Canberra, ACT 2612, Australia (e-mail: siqi.ma@adfa.edu.au).\protect
\IEEEcompsocthanksitem S. Camtepe and S. Nepal are with Data61, CSIRO, Sydney, NSW 1466, Australia (e-mail: sayit.camtepe@data61.csiro.au; surya.nepal@data61.csiro.au).}}

\markboth{IEEE Transactions on Multimedia,~Under review}%
{Shell \MakeLowercase{\textit{et al.}}: A Sample Article Using IEEEtran.cls for IEEE Journals}

\maketitle

\begin{abstract}
Deep neural networks (DNNs) have achieved state-of-the-art performance on face recognition (FR) tasks in the last decade. In real scenarios, the deployment of DNNs requires taking various face accessories into consideration, like glasses, hats, and masks. In the COVID-19 pandemic era, wearing face masks is one of the most effective ways to defend against the novel coronavirus. However, DNNs are known to be vulnerable to adversarial examples with a small but elaborated perturbation. Thus, a facial mask with adversarial perturbations may pose a great threat to the widely used deep learning-based FR models. In this paper, we consider a challenging adversarial setting: targeted attack against FR models. We propose a new stealthy physical masked FR attack via adversarial style optimization. Specifically, we train an adversarial style mask generator that hides adversarial perturbations inside style masks. Moreover, to ameliorate the phenomenon of sub-optimization with one fixed style, we propose to discover the optimal style given a target through style optimization in a continuous relaxation manner. We simultaneously optimize the generator and the style selection for generating strong and stealthy adversarial style masks. We evaluated the effectiveness and transferability of our proposed method via extensive white-box and black-box digital experiments. Furthermore, we also conducted physical attack experiments against local FR models and online platforms.
\end{abstract}


\section{Introduction}\label{sec:introduction}
\input{tex/1_introduction.tex}

\section{Related Work}\label{sec:related_work}
\input{tex/2_related_work.tex}

\section{Methodology}\label{sec:method}
\input{tex/3_method.tex}

\section{Experiments}\label{sec:experiments}
\input{tex/4_experiments.tex}

\section{Conclusion}\label{sec:conclusion}
\input{tex/5_conclusion.tex}


\bibliographystyle{IEEEtran}
\bibliography{main}

\newpage
\input{tex/7_bio.tex}

\vfill

\end{document}

%% file: tex/1_introduction.tex
\IEEEPARstart{I}{n} recent years, deep learning models have achieved great success in the face recognition (FR) task. A common deep FR model's working process is as follows: 1) utilizing convolutional neural networks (CNNs) to learn the face features of input face images; 2) constructing efficient loss function (like contrastive loss \cite{Hadsell2006,Sun2014}, triplet loss \cite{Schroff2015} and center loss \cite{Wen2016}) for distinguishing the face features; 3) using the well-trained models by 1) and 2) as well as some verification face images to define the distance threshold of different identities; 4) recognizing whether the two testing face images are the same person. Such an FR model and its improved versions \cite{Schroff2015,Wen2016,Liu2017,Wang2018,Deng2019,Sun2020b} can achieve very high accuracy on popular face image datasets. Therefore, these FR models are widely used in daily work and life. Many companies use these models to differentiate their employees and outsiders. Besides, lots of public places deploy them to find out specific persons. These high-accuracy FR models improve our life efficiency and enhance public security.

Since the end of 2019, a novel coronavirus (COVID-19 coronavirus) has been sweeping the world. Its effect on every aspect of our lives and work will continue in the near future. A simple and effective way to protect us from this virus is to wear face masks when we go into public places. However, the widely used face masks pose a threat to FR models: diverse face masks reduce the recognition rate; and, worse still, attackers deliberately elaborate some adversarial face masks to deceive FR models.

Adversarial vulnerability is a prevalent problem in the computer vision field: attackers elaborate examples with adversarial perturbations that can mislead the deep learning models \cite{Szegedy2014,Goodfellow2014,Amini2020,Zhong2022a}. Plenty of existing works have shown that adversarial examples (though with very small perturbations) can completely destroy the deep learning models \cite{Szegedy2014,Goodfellow2014}. Particularly, adversarial attacks are an enormous menace to FR models: black-box adversarial examples were proposed against FR models in \cite{Dong2019,Zhong2021}; adversarial face accessories were applied to attacking FR models (like glass, hat, and makeup) in \cite{Sharif2016,Komkov2020,Guetta2021}. However, most existing attacks achieve stealthiness by either crafting small perturbations or modifying semantic attributes of victim images. We wonder whether we can generate targeted stealthy adversarial perturbations against FR models.

In this paper, we propose a novel stealthy adversarial style mask to perform targeted attack against FR models in both digital and physical settings. Concretely, we train an adversarial style mask generator that hides adversarial perturbations inside the style masks. However, a fixed style usually incurs sub-optimal performance. To alleviate the this phenomenon, we introduce a continuous relaxation on style via a weighted sum of style set to discover the optimal style for a particular target. We optimize the style selection and the generator simultaneously for generating stealthy adversarial style masks (SASMask). Next, the SASMask is used to attack FR models in the digital setting and the physical setting. To better maintain the aggressivity in the physical world, we also adopt the adaptation mapping/transformation to robustize our SASMask. Before ending this section, we list the main contributions of this paper below:
\begin{itemize}
    \item We propose a novel stealthy adversarial style mask that hides adversarial perturbations inside the diverse style face masks via adversarial style optimization to attack FR models.
    \item We design to simultaneously train the adversarial style mask (SASMask) generator and optimize the style selection to alleviate the local minima problem and generate stronger adversarial masks.
    \item Extensive digital and physical experiments are conducted to demonstrate that the effectiveness of our proposed SASMask to deceive FR models.
\end{itemize}



%% file: tex/2_related_work.tex
\subsection{Face Recognition}
In recent decades, deep FR models has achieved great progress \cite{Ding2015,Schroff2015,Wen2016,Liu2017,Deng2019,Zhong2022a}. Existing deep FR models mostly utilize the idea of mapping a pair of face images to a distance proposed in \cite{Hadsell2005}. In 2015, Schroff {\it et al.} \cite{Schroff2015} proposed the effective triplet loss, which minimizes the distance between an anchor and a positive example and maximizes the distance between an anchor and a negative example until the margin is satisfied. Later, further works \cite{Wen2016,Liu2017,Wang2018,Deng2019} were proposed to improve the metric learning: Wen {\it et al.} \cite{Wen2016} proposed the center loss to learn a center for deep features of each class and penalize the distances between the deep features and their corresponding class centers; Liu {\it et al.} \cite{Liu2017} proposed the angular softmax loss that enables to learn angularly discriminative features; Wang {\it et al.} \cite{Wang2018} proposed the large margin cosine loss, and Deng {\it et al.} \cite{Deng2019} improve it by adding the parameter $m$ to the inside of $\cos$ function.

\subsection{Adversarial Attack}
First introduced in \cite{Szegedy2014}, the adversarial attack is to generate samples (i.e., adversarial examples) with imperceptible perturbations that can mislead deep models. From the attack scenario, the adversarial attack can be utilized in a digital setting and in a physical setting.

\noindent\textbf{Digital Attack.}
Numerous previous works show that deep models are vulnerable to adversarial examples in many fields, like classification \cite{MoosaviDezfooli2016,Cheng2021}, natural language processing \cite{Zhang2020}, object detection \cite{Xie2017}, speech recognition \cite{Qin2019}, etc. These attacks can be either restricted or unrestricted. The strong 1st-order Projected Gradient Descent (PGD) \cite{Madry2018} is a famous work of restricted attacks. Besides, unrestricted adversarial examples are also extensively studied, which try to alter some significant attributes or components of images, like color \cite{Hosseini2018, Laidlaw2019}, texture \cite{Zeng2019, Liu2019} as well as modifications of partial areas of images \cite{Brown2017, Hu2021}. Notwithstanding, such malicious samples are usually unnatural or distorted. Additionally, the unrestricted attacks cannot mount adversarial examples with complicated patterns, which limits their usage in practical situations. 
\noindent\textbf{Physical Attack.}
Kurakin {\it et al.} \cite{Kurakin2016} showed that the digital adversarial examples are still effective to subvert deep neural models by printing and recapturing them with a camera. However, the attack performance plummets because of viewpoint shifts, camera noise, or other physical transformations \cite{Athalye2018}. As such, more robust physical attacks demand larger perturbations, and adaptations towards physical transformations. Expectation Over Transformation (EoT) \cite{Athalye2018} is an effective technique to maintain the attack performance in the physical setting. Other robust physical attacks include adversarial patch (AdvPatch) \cite{Brown2017}, malicious graffiti \cite{Eykholt2018}, adversarial laser beams \cite{Duan2021}, adversarial clothes \cite{Hu2022}, adversarial shadows \cite{Zhong2022} and so forth. Most of these physical attacks generate large perturbations to increase the attack strength, which inevitably accounts for large and unnatural distortions of original images. Moreover, widely-used FR models attracts much attention to revealing the potential threats against them. Generally, such threats are divided into the dodging (untargeted) attack and the impersonation (targeted) attack. The dodging attack tries decrease the similarity of same-identity pairs, while the impersonation attack aims to increase the similarity between the test face and the target identity, which is more difficult than the untargeted attack technically. Thus, we mainly focus on the impersonation face attack in this paper. There are some related works of this task: adversarial glasses \cite{Sharif2016}, adversarial hat \cite{Komkov2020}, eye patch \cite{Xiao2021}, adversarial makeup \cite{Yin2021}. Besides, Zolfi {\it et al.} \cite{Zolfi2022} proposed an adversarial mask (AdvMask) that is quite related to our work. They chose a basic pattern and optimized it, which starts training the adversarial mask from a manually selected pattern, while our method not only optimizes the perturbation generation but also optimizes the style weights for selecting optimal basic patterns. Besides, AdvMask only uses the total variation loss as the constraint loss, which is not enough to constrain the stealthiness of the generated perturbations, whereas we propose explicit losses to constrain the magnitude of perturbations.

\subsection{Style Transfer}
Style transfer is to transfer the texture of a source image to a target image with retaining the content of the source image. With the help of deep neural networks, neural methods \cite{Gatys2016,Johnson2016,Dumoulin2017} achieved remarkable progress for the style transfer task, where the content and style features are learned by CNNs. Then, the style features of the style image are combined into the target image to realize style transfer. Recently, a few works \cite{Duan2020,Cao2022} generated adversarial examples with style transfer. In this paper, we hide adversarial perturbations in the masks to generate natural physical attacks against FR models. Different from existing works, we propose the adversarial attack in multiple styles and optimize the combination weights of these styles.



%% file: tex/3_method.tex
\begin{figure}[t]
\centering
\includegraphics[width=0.95\linewidth]{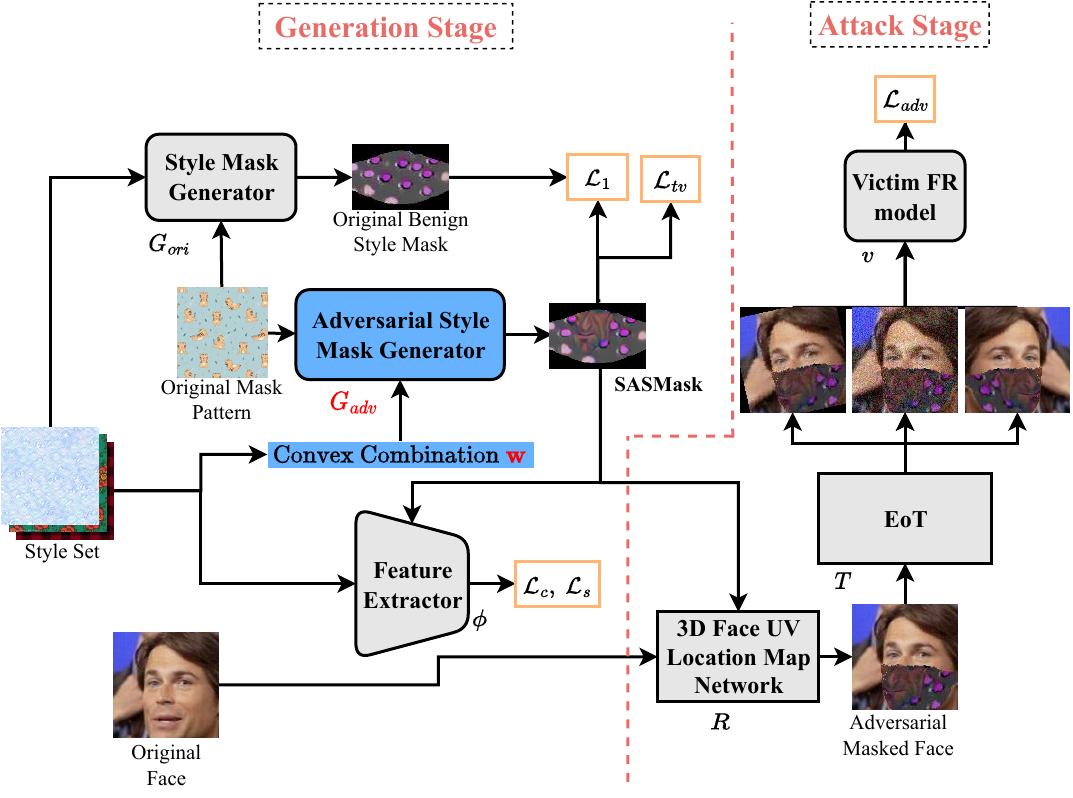}
\caption{Overview of our proposed method, which is divided into two stages: the generation stage; the attack stage. Parameters and networks that need to be optimized are emphasized in red bold font.}
\label{fig:method}
\vspace{-1em}
\end{figure}

The attack success rate (ASR) of adversarial attack is always regarded as one of the most important criteria, however, the naturalness of generated physical adversarial examples in real scenarios becomes another necessary factor in the evaluation \cite{Duan2020,Cao2022}. In this work, we focus on the targeted adversarial attack on physical masks via different styles. Previous style-transfer based physical attacks aims at untargeted attacks with a specific style. However, these techniques cannot be directly applied in our setting since we consider a more complex targeted attack. With a fixed style, it could be difficult for attackers to generate targeted physical adversarial masks with a better trade-offs between ASR and naturalness. We argue that different styles lead to different trade-offs for a target. Thus, instead of utilizing a fixed style for style-transfer attack which could result in the sub-optimal solutions, we propose to search the optimal style for each target in adversarial mask generation.

\subsection{Problem Formulation}\label{sec:problem_formulation}
Given a face image $x\in\mathbb{R}^m$ with its ground truth identity $y$, a face recognition model $v$ maps image to a feature embedding $e$, as well as a predicted identity. Given a target identity $y_t \neq y$, the goal of targeted adversarial attack on facial mask is to discover a natural mask with perturbation $M^{adv}$ which fools the model $v$ to predict the target identity. Formally, the adversarial mask is defined as
\begin{equation}\label{eqn:adv_attack}
\begin{aligned}
    & \min_{M^{adv}}~\mathcal{L}_{FR}(v(R(x,M^{adv})),e_t), \\
    & \text{where } e_t = v(x_t),\\
\end{aligned}
\end{equation}
where $\mathcal{L}_{FR}$ is the face recognition loss, $e_t$ denotes the feature embedding of target $y_t$, and $R(x,M^{adv})$ is the overlay operation that overlays adversarial mask $M^{adv}$ on face $x$. A traditional solution to Eq. \ref{eqn:adv_attack} can be adversarial attack via an additive noise as $M^{adv} = M + \delta$ where $M$ is the original mask and $\delta$ denotes the perturbation. Although the masks generated by adversarial attacks can be aggressive, those masks are always far away from those natural ones in the real world, which makes the attack perceptible in practise, such as PGD and AdvPatch \cite{Madry2018,Brown2017}. 

\subsection{Stealthy Adversarial Style Mask}\label{sec:method_1}
Instead of directly applying adversarial attacks, we propose to perform incorporate attack into a customized style. Specifically, we utilize the style transfer techniques to generate stealthy adversarial masks. Given the content of the original mask pattern $c$ and the style of predefined style image $s$, the adversarial mask $M^{adv}$ in Eq. \ref{eqn:adv_attack} can be generated via a style-transferred manner as
\begin{equation}\label{eqn:adv_mask}
  M^{adv}=G_{adv}(c,s).
\end{equation}
where $G_{adv}$ denotes the adversarial style transfer generator. Similarly, we denote the one without attack as $G_{ori}$. In order to guarantee stealthiness of $M^{adv}$, we include $l_1$-norm loss $\mathcal{L}_1$ to reduce the perturbation magnitude of image pixels, a total variation (TV) loss $\mathcal{L}_{tv}$ to reduce the variance of image pixels, a content loss $\mathcal{L}_c$ to remain the content of the source pattern and a style loss $\mathcal{L}_s$ to generate the customized style. Formally, given a pretrained feature extractor $\phi(\cdot)$, these loss terms can be formulated as
\begin{equation} \label{eqn:style_loss}
\begin{aligned}
    & \mathcal{L}_{1}= |M^{adv}-M|=|G_{adv}(c,s)-G_{ori}(c,s)|,\\
    & \mathcal{L}_{tv}=\sum_{i,j}\sqrt{(M^{adv}_{i,j}-M^{adv}_{i+1,j})^2+(M^{adv}_{i,j}-M^{adv}_{i,j+1})^2},\\
    & \mathcal{L}_c=\sum_{i\in l_{C}}\left \| \phi_i(M^{adv})-\phi_i(c) \right \|_2^2,\\
	& \mathcal{L}_s=\sum_{j\in l_{S}}\left \| \mathcal{G}(\phi_j(M^{adv}))-\mathcal{G}(\phi_j(s)) \right \|_2^2,\\
\end{aligned}
\end{equation}
where $\phi_l(\cdot)$ is the output of the feature extractor at layer $l$, $l_{C}$ and $l_{S}$ are the sets of content and style layers; $\mathcal{G}(\phi_l(\cdot))$ is the Gram matrix associated with the output at layer $l$.
Thus Eq. \ref{eqn:adv_attack} can be reformulated as
\begin{equation}\label{eqn:total_loss}
\begin{aligned}
& \min_{G_{adv}}~\mathcal{L}(v,R,G_{adv},G_{ori},\phi;x,e_t,c,s),\\
&\text{where } \mathcal{L} = \mathcal{L}_{FR} + \lambda_1\mathcal{L}_1 + \lambda_{tv}\mathcal{L}_{tv} + \lambda_c\mathcal{L}_c + \lambda_s\mathcal{L}_s,\\
\end{aligned}
\end{equation}
where $\lambda_1$, $\lambda_{tv}$, $\lambda_c$, and $\lambda_s$ denote the scaling hyperparameters.

\noindent\textbf{Physical Adaptation.}
Since physical environments are not as ideal as digital conditions, it usually involves noises (like environmental noises and camera noises) as well as natural transformations (like viewpoint shifts and rotations) in a physical setting. Therefore, we need to use physical adaptations to handle such diverse conditions. Here, we use the end-to-end 3D face UV position map \cite{Feng2018} as the overlay operation $R(x,M^{adv})$ to overlay facial masks on faces for generating vivid and natural physically adaptive facial masks. Besides, we also utilize an auxiliary method similar to Expectation Over Transformation (EoT) \cite{Athalye2018}. In our setting, we aim to improve the robustness of SASMask towards noises, rotations, and viewpoint shifts. Here, we use a transformation function $T(\cdot)$ to transform the masked face data as $T(R(x,M^{adv}))$.

\begin{algorithm}[tbp]
\caption{Algorithm for SASMask}\label{alg:method}
\KwData{face images $x$, an original mask pattern $c$, style set $S$, trained style transfer network $G_{ori}$, targeted face image $x_t$, victim face recognition model $v$, training epoch $N$}
\KwResult{SASMask generator $G_{adv}$, optimized style weights $\mathbf{w}$}

Initialize $G^0_{adv}=G_{ori}$, $\mathbf{w}^0=[0]_K$,\;
Generate original style mask $G_{ori}(c,\Tilde{h}^0 \cdot S)$\;
Compute targeted embedding $e_t=v(x_t)$\;
\For{$n=1,\cdots,N$}{
    Get SASMask by $G_{adv}(c,\Tilde{h}^0 \cdot S)$\;
    Get the mask position map of SASMask by \cite{Feng2018} and overlay it on face images to obtain $R(x,G_{adv}(x,\Tilde{h}^0 \cdot S))$   \;
    Randomly transform the masked faces $T(R(x,G_{adv}(c,\Tilde{h}^0 \cdot S)))$ \;
    Calculate the loss in Eq. \ref{eqn:opt_relax}\;
    Separately compute generator gradient $\nabla_{G^n_{adv}}\mathcal{L}$ and weight gradient $\nabla_{\mathbf{w}^n}\mathcal{L}$\;
    Update $G^{n+1}_{adv}$ with $G^{n}_{adv}$ and $\nabla_{G^n_{adv}}\mathcal{L}$ by some optimizer (like ADAM \cite{Kingma2015})\;
    Update $\mathbf{w}^{n+1}$ with $\mathbf{w}^{n}$ and $\nabla_{\mathbf{w}^n}\mathcal{L}$ by some optimizer\;
}
\end{algorithm}

\subsection{Adversarial Style Optimization}

\begin{table}[tbp]
\centering
\caption{Architecture of (adversarial) style mask generator.}
\label{tab:style-network}
\resizebox{\linewidth}{!}{
\begin{tabular}{clcc}
\toprule[1.5pt]
\textbf{Index} & \multicolumn{1}{c}{\textbf{Operation}}                                                                                                                                       & \textbf{Input Size} & \textbf{Output Size} \\
\midrule[1pt]
1              & Conv(in=3,out=32,k=9,p=(4,4,4,4),s=1)+InstanceNorm+ReLU                                                                                                                      & (3,112,112)         & (32,112,112)         \\
\midrule[0.5pt]
2              & Conv(in=32,out=64,k=3,p=(0,1,0,1),s=2)+InstanceNorm+ReLU                                                                                                                     & (32,112,112)        & (64,56,56)           \\
\midrule[0.5pt]
3              & Conv(in=64,out=128,k=3,p=(0,1,0,1),s=2)+InstanceNorm+ReLU                                                                                                                    & (64,56,56)          & (128,28,28)          \\
\midrule[0.5pt]
4              & $\left[
\begin{tabular}[c]{@{}l@{}}Conv(in=128,out=128,k=3,p=(1,1,1,1),s=1)+InstanceNorm+ReLU\\ Conv(in=128,out=128,k=3,p=(1,1,1,1),s=1)+InstanceNorm+Residual Addition\end{tabular}\right]$$\times 5$ & (128,28,28)         & (128,28,28)          \\
\midrule[0.5pt]
5              & \begin{tabular}[c]{@{}l@{}}Interpolate(scale\_factor=2)\\ Conv(in=128,out=64,k=3,p=(1,1,1,1),s=1)+InstanceNorm+ReLU\end{tabular}                                             & (128,28,28)         & (64,56,56)           \\
\midrule[0.5pt]
6              & \begin{tabular}[c]{@{}l@{}}Interpolate(scale\_factor=2)\\ Conv(in=64,out=32,k=3,p=(1,1,1,1),s=1)+InstanceNorm+ReLU\end{tabular}                                              & (64,56,56)          & (32,112,112)         \\
\midrule[0.5pt]
7              & Conv(in=32,out=3,k=9,p=(4,4,4,4),s=1)+InstanceNorm+Sigmoid                                                                                                                   & (32,112,112)        & (3,112,112) \\
\bottomrule[1.5pt]
\end{tabular}
}
\vspace{-1em}
\end{table}

Note that the stealthy adversarial style masks generated by Eq. \ref{eqn:total_loss} could achieve sub-optimal adversarial strength, because a fixed style is not always a good initial point for different targets. As shown in Fig. \ref{fig:sty_ab}, when we randomly sample one style from the style set, the optimization stops at different points in which the attack performances vary greatly. We mainly attribute this instability to sub-optimal solution via a fixed style. Thus, in order to perform stable effective targeted adversarial style masks, styles are incorporated into the optimization. Specifically, we propose to optimize the convex combination of adversarial styles from a style set. Given a style set $S$ that contains $K$ style candidates, we aim at discovering a superior style which achieves better trade-offs between attack success rate and stealthiness than other potential styles. The objective in Eq. \ref{eqn:total_loss} becomes a bi-level optimization which can be reformulated as
\begin{equation} \label{eqn:bi_level_obj}
\begin{aligned}
    \min_{G_{adv}}~ & \mathcal{L}(v,R,G_{adv},G_{ori},\phi;x,e_t,c,s^*),\\
    \textbf{s.t. } &  s^* = \argmin_{s \in S} \mathcal{L}(v,R,G_{adv},G_{ori},\phi;x,e_t,c,s)\\
\end{aligned}
\end{equation}
Because style selection is a discrete problem as shown in Eq. \ref{eqn:bi_level_obj}, it is difficult to optimize it directly. Hence, we propose to conduct continuous relaxation on $s$ via the incorporation of style weight as:
\begin{equation}\label{eqn:style_select}
s = \mathbf{w}\cdot S=\sum^{K}_kw_k S_k.
\end{equation}
where $\mathbf{w}=[w_1,\cdots,w_K]$ is the weight vector of $K$ styles so that the style $s$ becomes a weighted sum of style set $S$. During the inference stage, the style with maximum $\mathbf{w}$ is selected as the optimal style for adversarial style attack as
\begin{equation} \label{eqn:style_argmax}
\begin{aligned}
    & s^* = h \cdot S, \text{ where } h = \text{one\_hot} (\argmax_{k} w_k),\\
\end{aligned}
\end{equation}
where $h$ denotes the one-hot distribution. Since Eq. \ref{eqn:style_argmax} is not differentiable in training phase, we propose to use the softmax with a temperature parameter to approximate the one-hot distribution $h$ in a differentiable manner as
\begin{figure}[t]
\centering
\includegraphics[width=0.9\linewidth]{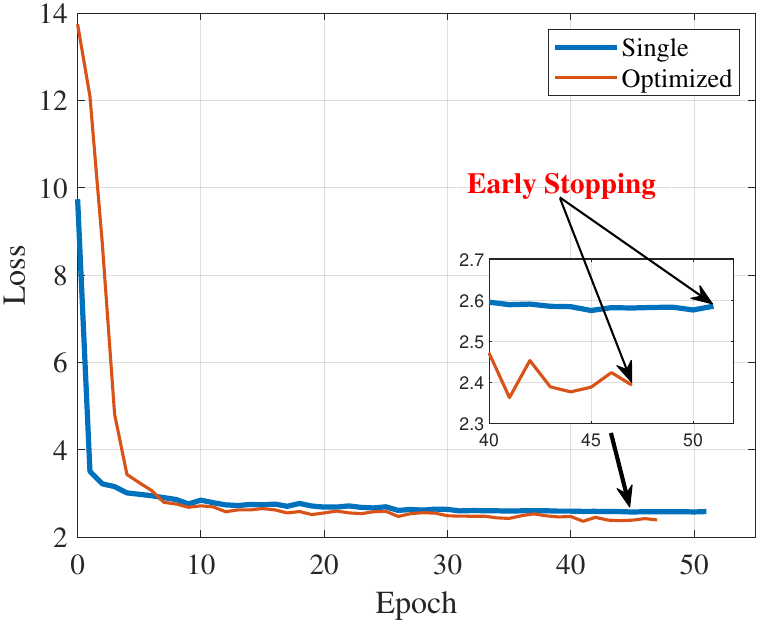}
\caption{Convergence curves of optimized and single adversarial style masks.}
\label{fig:converge}
\vspace{-1em}
\end{figure}
\begin{table}[t]
\centering
\caption{ASR, SSIM and Trade-off with hyperparameter tuning.}
\label{tab:tuning}
\resizebox{\linewidth}{!}{
\begin{tabular}{ccccccc}
    \toprule[1pt]
    $\lambda_1$ & $\lambda_{tv}$ & $\lambda_{c}$   & $\lambda_{s}$  & ASR             & SSIM            & Trade-off      \\
    \midrule[0.5pt]
    100   & 10    & 0.001  & 1000  & 98.8\%          & 0.9130          & \textbf{1.901} \\
    \midrule[0.5pt]
    10    & 10    & 0.001  & 1000  & \textbf{99.1\%} & 0.8732 & 1.864          \\
    1000  & 10    & 0.001  & 1000  & 0.0\%           & \textbf{0.9265} & 0.927          \\
    \midrule[0.5pt]
    100   & 1     & 0.001  & 1000  & 98.8\% & 0.9066          & 1.895          \\
    100   & 100   & 0.001  & 1000  & 0.0\%  & 0.9220 & 0.922          \\
    \midrule[0.5pt]
    100   & 10    & 0.0001 & 1000  & 98.9\%          & 0.9037          & 1.893          \\
    100   & 10    & 0.01   & 1000  & 97.6\% & 0.9146 & 1.891          \\
    \midrule[0.5pt]
    100   & 10    & 0.001  & 100   & 98.3\% & 0.9134 & 1.896          \\
    100   & 10    & 0.001  & 10000 & 98.4\% & 0.8993 & 1.883   \\
    \bottomrule[1pt]
\end{tabular}
}
\vspace{-1em}
\end{table}


\begin{equation}\label{eqn:style_weights}
\Tilde{h}_{i}(w_i) = \frac{\text{exp}(w_i)/\tau}{\sum^K_{j=1} \text{exp}(w_j)/\tau},
\end{equation}
where $\tau> 0$ is the temperature parameter, used for controlling the concentration of style selection. Through setting $\tau$ to a smaller value during the training phase (e.g., 0.1), $\Tilde{h}_{i}$ is close to one-hot distribution $h$, which alleviates the gap between training and inference phases. The objective in Eq. \ref{eqn:bi_level_obj} can be reformulated as
\begin{equation} \label{eqn:opt_relax}
\begin{aligned}
    \min_{G_{adv}}~ & \mathcal{L}(v,R,G_{adv},G_{ori},\phi;x,e_t,c,s^*),\\
    \min_{\mathbf{w}}~ & \mathcal{L}(v,R,G_{adv},G_{ori},\phi;x,e_t,c,\mathbf{w},S), \\
    \textbf{s.t. }&  s^* = \Tilde{h}(\mathbf{w}) \cdot S.\\
\end{aligned}
\end{equation}

During the inference phase, $\Tilde{h}$ is replaced by the one-hot distribution $h$ via argmax on $\mathbf{w}$ to select the optimal style $s^*$ given a target. Thus, with the involvement of style weights, an optimal style can be discovered via our framework for each target instead of the sub-optimal solutions with a fixed style in previous works. We further summarize our proposed algorithm in Fig. \ref{fig:method} and Algorithm \ref{alg:method}.

%% file: tex/4_experiments.tex
\begin{table*}[tbp]
\centering
\caption{ASR and SSIM of different attack methods with different hyperparameters of four different targets. Here, we study the ResNet-50 backbone and the ArcFace head as the victim FR model. The best results are stressed in \textbf{bold}.}
\label{tab:mul_targets}
\resizebox{\linewidth}{!}{
\begin{tabular}{cccccccccccc}
\toprule[1.5pt]
\textbf{Target}                    & \textbf{Attack   Method} & Original   Face & Rand   & PGD-64 & PGD-80 & PGD-128 & AdvPatch-20 & AdvPatch-24 & AdvPatch-28 & AdvMask & \textbf{Ours} \\
\midrule[1pt]
\multirow{2}{*}{Vivica   Fox}      & ASR                      & 0.1\%           & 0.3\%  & 48.9\% & 49.6\% & 51.1\%  & 3.9\%       & 4.9\%       & 5.1\%       & 98.6\%  & \textbf{98.8\%}         \\
                                   & SSIM                     & -               & 0.7166 & 0.8998 & 0.8858 & 0.8545  & 0.8265      & 0.7964      & 0.7590      & 0.6614  & \textbf{0.9130}         \\
\midrule[1pt]
\multirow{2}{*}{Patricia   Hearst} & ASR                      & 0.7\%           & 2.7\%  & 9.1\%  & 10.1\% & 12.3\%  & 6.6\%       & 6.8\%       & 7.9\%       & 81.7\%  & \textbf{85.3\%}         \\
                                   & SSIM                     & -               & 0.7290 & 0.9001 & 0.8979 & 0.8963  & 0.7160      & 0.6580      & 0.5868      & 0.5926  & \textbf{0.9122}         \\
\midrule[1pt]
\multirow{2}{*}{Aaron   Eckhart}   & ASR                      & 0.0\%           & 0.1\%  & 3.6\%  & 8.4\%  & 30.5\%  & 0.0\%       & 0.0\%       & 0.1\%       & 89.3\%  & \textbf{91.7\%}         \\
                                   & SSIM                     & -               & 0.7418 & 0.8541 & 0.8445 & 0.7871  & 0.8359      & 0.8089      & 0.7920      & 0.7683  & \textbf{0.8844}         \\
\midrule[1pt]
\multirow{2}{*}{Steve   Park}      & ASR                      & 0.9\%           & 1.3\%  & 9.4\%  & 8.2\%  & 10.6\%  & 3.3\%       & 4.6\%       & 5.3\%       & 78.5\%  & \textbf{84.4\%}         \\
                                   & SSIM                     & -               & 0.7591 & 0.9065 & 0.8969 & 0.8887  & 0.7852      & 0.7026      & 0.6222      & 0.6082  & \textbf{0.9181} \\
\bottomrule[1.5pt]
\end{tabular}
}
\vspace{-1em}
\end{table*}
\begin{table*}[tbp]
\centering
\caption{Multi-target attack experimental results.}
\label{tab:multi-target}
\resizebox{\linewidth}{!}{
\begin{tabular}{ccccccccccccc}
\toprule[1.5pt]
\# of Targets & Metric & No Mask & Ori Mask & Rand & PGD-64 & PGD-80 & PGD-128 & AdvPatch-20 & AdvPatch-24 & AdvPatch-28 & AdvMask & Ours \\
\midrule[1pt]
\multirow{2}{*}{5}  & ASR  & 4.6\% & 6.3\% & 4.2\%  & 39.9\% & 48.1\% & 52.4\% & 7.2\%  & 8.6\%  & 8.1\%  & 88.4\% & 90.6\% \\
                    & SSIM & -     & -     & 0.7205 & 0.8966 & 0.8831 & 0.8430 & 0.8320 & 0.8031 & 0.7718 & 0.6698 & 0.9177 \\
\multirow{2}{*}{10} & ASR  & 6.9\% & 6.0\% & 5.2\%  & 41.9\% & 47.7\% & 47.6\% & 7.9\%  & 8.5\%  & 9.0\%  & 87.2\% & 89.8\% \\
                    & SSIM & -     & -     & 0.7205 & 0.8951 & 0.8810 & 0.8432 & 0.8345 & 0.8044 & 0.7749 & 0.6702 & 0.9150 \\
\multirow{2}{*}{15} & ASR  & 7.8\% & 4.3\% & 5.3\%  & 47.9\% & 53.0\% & 53.8\% & 8.1\%  & 8.3\%  & 8.8\%  & 86.5\% & 90.1\% \\
                    & SSIM & -     & -     & 0.7203 & 0.8948 & 0.8805 & 0.8431 & 0.8380 & 0.8066 & 0.7748 & 0.6703 & 0.9153 \\
\multirow{2}{*}{20} & ASR  & 7.7\% & 3.4\% & 4.4\%  & 46.4\% & 50.6\% & 53.0\% & 6.8\%  & 7.0\%  & 7.4\%  & 85.8\% & 90.5\% \\
                    & SSIM & -     & -     & 0.7207 & 0.8948 & 0.8808 & 0.8431 & 0.8392 & 0.8078 & 0.7782 & 0.6709 & 0.9155 \\
\multirow{2}{*}{30} & ASR  & 7.2\% & 3.3\% & 3.3\%  & 42.1\% & 46.2\% & 49.4\% & 5.3\%  & 5.4\%  & 5.8\%  & 80.6\% & 90.8\% \\
                    & SSIM &       & -     & 0.7206 & 0.8948 & 0.8812 & 0.8433 & 0.8409 & 0.8094 & 0.7786 & 0.6668 & 0.9162 \\
\multirow{2}{*}{40} & ASR  & 6.6\% & 3.6\% & 2.9\%  & 39.6\% & 43.1\% & 45.7\% & 4.7\%  & 4.7\%  & 5.1\%  & 76.2\% & 90.0\% \\
                    & SSIM & -     & -     & 0.7203 & 0.8955 & 0.8821 & 0.8443 & 0.8414 & 0.8093 & 0.7781 & 0.6663 & 0.9167 \\
\multirow{2}{*}{50} & ASR  & 6.1\% & 3.5\% & 2.4\%  & 36.4\% & 39.2\% & 41.6\% & 4.2\%  & 4.2\%  & 4.6\%  & 69.8\% & 89.8\% \\
                    & SSIM & -     & -     & 0.7201 & 0.8956 & 0.8822 & 0.8442 & 0.8410 & 0.8094 & 0.7779 & 0.6670 & 0.9170 \\
\bottomrule[1.5pt]
\end{tabular}
}
\vspace{-1em}
\end{table*}
\begin{figure*}[tbp]
\centering
\subfloat[]{\includegraphics[width=0.49\linewidth]{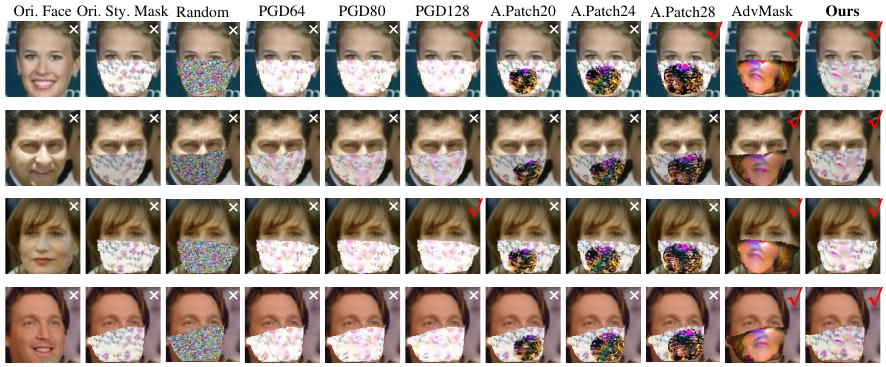}%
\label{fig:one_target}}
\hfil
\subfloat[]{\includegraphics[width=0.49\linewidth]{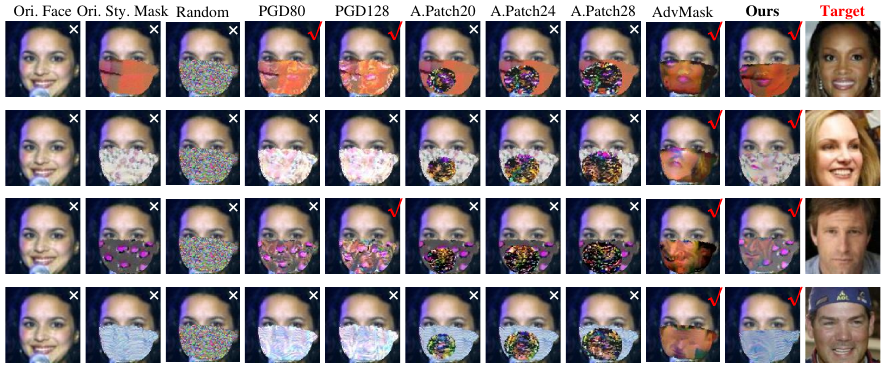}%
\label{fig:mul_targets}}
\caption{Adversarial masked face images generated by different attack methods, where images with red check marks on the right top denote successful attacks against FR models, while white crosses denote failed attacks. Note that masks by AdvMask and our method both show strong attack performances, while our masks provide much better stealthiness (much higher SSIM) than AdvMask. (a) Adversarial examples for the target Patricia Hearst (cf. the image in the second row, the right-most column in Fig. \ref{fig:mul_targets}). (b) Adversarial examples for four targets: Vivica Fox, Patricia Hearst, Aaron Eckhart, Steve Park (from top to bottom).}
\label{fig:digital_examples}
\vspace{-1em}
\end{figure*}

In this section, we evaluated the effectiveness of SASMask in the digital setting and the physical setting. Specifically, we first outlined the experimental setup, and then we displayed the white-box and black-box experimental results in the digital setting, including multiple targets, cross-backbone, cross-head, and cross-dataset experiments. Besides, we analyzed SASMask via the ablation study for adversarial style optimization and loss ablation. Eventually, we also conducted physical experiments: offline printed experiments and online platform attack experiments.

\subsection{Experimental Setup}
\paragraph{FR model and mask generator}
To fully evaluate our proposed method, we utilized five backbones and five state-of-the-art metric learning heads to train our SASMask generator and tested the trained model in the white-box and black-box settings. The five backbones are ResNet (-34, -50, -101) \cite{He2016}, MobileNet \cite{Chen2018a} and GhostNet \cite{Han2020}. Additionally, the heads are ArcFace \cite{Deng2019}, CosFace \cite{Wang2018}, CircleLoss \cite{Sun2020a}, CurricularFace \cite{Huang2020}, MagFace \cite{Meng2021}. For the (adversarial) style mask generator, we design a simplified U-net \cite{Ronneberger2015} like network to generate (adversarial) style masks, whose architecture is shown in Table \ref{tab:style-network}.  

\paragraph{Dataset}
We used four popular facial image datasets in our experiments: LFW \cite{Huang2008}, VGGFace2 \cite{Cao2018}, AgeDB \cite{Moschoglou2017}, CFP \cite{Sengupta2016}. We followed the setup of existing works, e.g., \cite{Dong2019,Yin2021,Zolfi2022}. We randomly chose fifty identities from the LFW dataset as our targets for comprehensively evaluating the performance of our methods. In the training stage, we randomly selected 400 images of different identities (different from targeted identities) from the LFW dataset. In the testing stage, we randomly chose another 1000 images (different from targeted identities and training images) from the LFW dataset. Besides, we also randomly selected 1000 facial images from the other three datasets (VGGFace2, AgeDB, and CFP) for testing the transferability to other datasets. For the training stage, subset data is enough, because there is little improvement with the full set, which needs much more computational resources.

\paragraph{Baseline attack}
We compared our proposed method with several existing representative adversarial methods: Projected Gradient Descent (PGD) \cite{Madry2018}, Adversarial Patch (AdvPatch) \cite{Brown2017}, and AdvMask \cite{Zolfi2022}. PGD is the strongest first-order digital attack; AdvPatch is one of the typical unrestricted attacks that can be directly used in the physical setting; AdvMask is a recently proposed adversarial mask. In addition, we also generated the randomly perturbed face mask as another baseline attack method. In addition to the above mask region-based comparisons, we also compared our method with other face attack methods: PGD on face (PGDFace) \cite{Madry2018}, AdvPatch (PatchFace) on face \cite{Xiao2021}, AdvGlass \cite{Sharif2016}, AdvHat \cite{Komkov2020}, AdvMakeup \cite{Yin2021}. For all the comparison methods, we follow their official experimental settings.

\begin{table}[tbp]
\centering
\scriptsize
\caption{Average transfer ASR of adversarial masked face examples generated by different attack methods with different backbones.}
\label{tab:transfer-backbone}
\resizebox{\linewidth}{!}{
\begin{tabular}{lccccc}
\toprule[1.5pt]
           & Rand  & PGD    & AdvPatch & AdvMask & \textbf{Ours}   \\
\midrule[1pt]
ResNet-34 \cite{He2016}  & 1.1\% & 4.1\%  & 2.2\%    & 76.3\%  & \textbf{87.5\%} \\
ResNet-50 \cite{He2016} & 0.3\% & 13.1\% & 1.5\%    & 61.9\%  & \textbf{74.1\%} \\
ResNet-101 \cite{He2016} & 0.6\% & 20.0\% & 1.8\%    & 57.3\%  & \textbf{71.5\%} \\
MobileNet \cite{Chen2018a}  & 0.7\% & 2.0\%  & 2.0\%    & 41.2\%  & \textbf{43.4\%} \\
GhostNet \cite{Han2020}  & 0.2\% & 5.8\%  & 1.4\%    & 68.0\%  & \textbf{72.7\%} \\
\bottomrule[1.5pt]
\end{tabular}
}
\end{table}
\begin{table}[tbp]
\centering
\scriptsize
\caption{Average transfer ASR of adversarial masked face examples generated by different attack methods with different heads.}
\label{tab:transfer-head}
\resizebox{\linewidth}{!}{
\begin{tabular}{lccccc}
\toprule[1pt] 
               & Rand & PGD    & AdvPatch & AdvMask & \textbf{Ours}   \\
\midrule[1pt]
ArcFace \cite{Deng2019}& 0.3\%  & 2.0\%  & 0.4\%    & 43.2\%  & \textbf{51.7\%} \\
CosFace  \cite{Wang2018}& 0.0\%  & 18.7\% & 0.4\%    & 68.2\%  & \textbf{74.0\%} \\
CircleLoss \cite{Sun2020a}& 0.1\%  & 11.1\% & 0.8\%    & 70.5\%  & \textbf{76.5\%} \\
CurricularFace \cite{Huang2020}& 0.1\%  & 7.6\%  & 0.6\%    & \textbf{75.1\%}  & 72.4\% \\
MagFace \cite{Meng2021}& 0.7\%  & 8.5\%  & 0.5\%    & 68.2\%  & \textbf{71.2\%} \\
\bottomrule[1pt]
\end{tabular}
}
\end{table}
\paragraph{Metric}
To quantify the attack performance of our method and the comparison methods, we used the \emph{attack success rate} (ASR) as our attack metric, which is here defined as 
\begin{equation}\label{eqn:asr}
\text{ASR}=\frac{\sum \mathds{1}(d(\text{Test Face, Target})<\kappa)}{\text{\# of All Test Pairs}}\times 100\%,
\end{equation}
where $d(\cdot,\cdot)$ is the distance metric; $\kappa$ is the distance threshold and we used the $K$-fold cross validation of the full test set to calculate it. Moreover, we also utilized the \emph{structure similarity} (SSIM) \cite{Wang2004} as our mask quality metric, whose definition is given by
\begin{equation}\label{eqn:ssim}
  \text{SSIM}(a,b)=\frac{(2\mu_a\mu_b+C_1)\cdot(2\sigma_{ab}+C_2)}{(\mu_a^2+\mu_b^2+C_1)\cdot(\sigma_a^2+\sigma_b^2+C_2)},
\end{equation}
where $a$ and $b$ are two masked facial images; $\mu_a$ (or $\mu_b$) is the image mean; $\sigma_a$ (or $\sigma_b$) is the image standard deviation; $\sigma_{ab}$ denotes the covariance between $a$ and $b$; $C_1$ and $C_2$ are two constants ($C_1=0.0001$ and $C_2=0.0009$ in our experiments). Hereof, we computed the SSIM between the original style masked face image and the attacked masked image to evaluate the stealthiness of the deployed attack method, where larger SSIM means better stealthiness. Note that we used the value of SSIM to quantitationally calculate the metric of stealthiness (i.e., naturalness, the quality or state of being natural).
\begin{table}[tbp]
\centering
\scriptsize
\caption{Transfer ASR and SSIM of adversarial masked face images generated by different attack methods from one dataset to the other three datasets.}
\label{tab:dataset_trans}
\resizebox{\linewidth}{!}{
\begin{tabular}{lccccc}
\toprule[1pt]
\multicolumn{2}{c}{Dataset} & {LFW}    & {VGGFace2} & {AgeDB}   & {CFP}    \\
\midrule[0.5pt]
\multirow{2}{*}{Rand}   & ASR    & 0.3\%           & 3.1\%             & 1.2\%            & 0.5\%           \\
                                   & SSIM    & 0.7166          & 0.7380             & 0.7550            & 0.7910           \\
\midrule[0.5pt]
\multirow{2}{*}{PGD}      & ASR    & 51.1\%          & 82.8\%            & 81.1\%           & 69.4\%          \\
                                   & {SSIM}    & 0.8545          & 0.8628            & 0.8724           & 0.8917          \\
\midrule[0.5pt]
\multirow{2}{*}{AdvPatch} & {ASR}     & 5.1\%           & 10.5\%            & 5.6\%            & 5.4\%           \\
                                   & {SSIM}    & 0.7590           & 0.7623            & 0.7827           & 0.8080           \\
\midrule[0.5pt]
\multirow{2}{*}{AdvMask}  & {ASR}     & 98.6\%          & \textbf{100.0\%}  & \textbf{100.0\%} & \textbf{99.9\%} \\
                                   & {SSIM}    & 0.6614          & 0.6620             & 0.6792           & 0.6931          \\
\midrule[0.5pt]
\multirow{2}{*}{\textbf{Ours}}     & {ASR}     & \textbf{98.8\%} & 99.9\%            & \textbf{100.0\%} & \textbf{99.9\%} \\
                                   & {SSIM}    & \textbf{0.9130}  & \textbf{0.9154}   & \textbf{0.9197}  & \textbf{0.9334}\\
\bottomrule[1pt]
\end{tabular}
}
\vspace{-1em}
\end{table}
\begin{table}[tbp]
\centering
\scriptsize
\caption{Comparison with face attack methods on LFW dataset.}
\label{tab:face_attack_comparison}
\resizebox{\linewidth}{!}{
\begin{tabular}{ccccccc}
\toprule[1pt]
Method & PGDFace    & AdvHat & AdvGlass & PatchFace & AdvMakeup & Ours   \\
\midrule[0.5pt]
ASR     & 14.6\% & 20.4\% & 12.4\%   & 10.3\%   & 22.0\%    & \textbf{43.4\%} \\
\bottomrule[1pt]
\end{tabular}
}
\vspace{-1em}
\end{table}
\paragraph{Implementation detail}
We first pre-trained a benign style transfer generator $G_{ori}$ on the Microsoft COCO dataset \cite{Lin2014}. Then, we used this generator to initialize the SASMask generator and fine-trained it with the Adam optimizer \cite{Kingma2015}. We set the initial model learning rate as $0.01$ and the weight learning rate as $0.01$. The values of $\lambda_{1}$, $\lambda_{tv}$, $\lambda_{c}$ and $\lambda_{s}$ were set to 100, 10, 0.001 and 1000, respectively. Note that we utilized the FR heads and the classification idea to training the FR backbones; in the testing stage, we only use the trained FR backbones to obtain the feature embeddings for verification. Besides, all the victim face recognition models are trained on the CASIA-WebFace dataset \cite{Yi2014}. The temperature $\tau$ was set to 0.1. For the input format, we set the size of input masked face images to be $[112,112]$, and set the size of masks to be $[60,112]$ in the bottom area of face images. We set the starting optimized points of different attack methods to be the original style masks. For the feature extractor, we use the VGG-16 architecture \cite{Simonyan2015}, where we use the ReLU outputs of layers 2, 4, 7 and 10 to compute the content and style losses. For the end-to-end 3D face UV location map network, we use the well-trained PRNet \cite{Feng2018}. For the batch normalization of (masked) facial images, we use the mean of $(0.5,0.5,0.5)$ and the standard deviation of $(0.5,0.5,0.5)$; for the batch normalization of VGG-16 inputs, we use the mean of $(0.485, 0.456, 0.406)$ and the standard deviation of $(0.229, 0.224, 0.225)$.

\paragraph{Hyperparameter tuning}
For the training stage, we first ran the model without optimization and obtain every loss, then initialize the hyperparameters to make them the same order of magnitude, and use the variable-controlling method to tune hyperparameters. Besides, we further used the early stopping trick \cite{Rice2020} that stops the training when there is no improvement after seven epochs for quicker convergence and better performance. To tune hyperparameter, we first run the model without optimization and get the value of every loss, then we initialize the hyperparameters to make each loss the same order of magnitude. At last, we used the variable-controlling approach to tune hyperparameters for better performance. Some funing results are shown in Table \ref{tab:tuning}, where the trade-off column denotes the ``sum'' of ASR and SSIM, which is used to describe the performance with consideration of attack and vision effect. Bigger trade-off value roughly means better performance. Besides, we use the Adam optimizer and the early stopping trick to train our models and it is easy to converge: normally, the training process takes about 50 epochs (about 3 hours) with a single GeForce RTX 2080 Ti GPU. The convergence curves are shown in Fig. \ref{fig:converge}.
\begin{table}[tbp]
\centering
\footnotesize
\caption{ASR and SSIM of adversarial masked face images by ablating different stealthy loss parts.}
\label{tab:loss_ab}
\resizebox{\linewidth}{!}{
\begin{threeparttable}
\begin{tabular}{ccccccc}
\toprule[1pt]
$\mathcal{L}_{adv}$ & $\mathcal{L}_{1}$ & $\mathcal{L}_{tv}$ & $\mathcal{L}_{c}$ & $\mathcal{L}_{s}$ & {ASR} & {SSIM} \\
\midrule[1pt]
\textbf{\checkmark} & \checkmark & \checkmark & \checkmark & \checkmark & 98.8\% & \textbf{0.9130} \\
\midrule[0.5pt]
\checkmark & - & \checkmark & \checkmark & \checkmark & 97.6\% & 0.8337 \\
\checkmark & \checkmark & - & \checkmark & \checkmark & 98.3\% & 0.8437 \\
\checkmark & \checkmark & \checkmark & - & \checkmark & 98.6\% & 0.8837 \\
\checkmark & \checkmark & \checkmark & \checkmark & - & 98.3\% & 0.8993 \\
\midrule[0.5pt]
\checkmark & - & - & \checkmark & \checkmark & 98.7\% & 0.7886 \\
\checkmark & \checkmark & \checkmark & - & - & 98.5\% & 0.8397 \\
\midrule[0.5pt]
\checkmark & - & - & - & - & \textbf{99.7\%}\tnote{*} & 0.6255 \\
 \bottomrule[1pt]
\end{tabular}
\begin{tablenotes}
\footnotesize
\item[*] ASR without any constraint loss is the ASR \textbf{upper bound} in our setting, which has an adverse side effect of very low SSIM.
\end{tablenotes}
\end{threeparttable}
}
\vspace{-1em}
\end{table}
\subsection{Digital Attack}
\subsubsection{White-Box Attack} 
We first evaluated our proposed method in a white-box setting. Specifically, we randomly selected four identities from the LFW dataset as the targets and trained corresponding SASMask generators. Here, we used the ResNet-50 backbone and the ArcFace head. Furthermore, we also tested the baseline attack method, i.e, random perturbations, PGD, AdvPatch, and AdvMask. For PGD attack methods, the maximum perturbations were set to be $64/255$, $80/255$ and $128/255$ with 40 iteration steps; For AdvPatch attack methods, the radii of the patches were set to be $20$, $24$, and $28$ (a larger radius would be beyond the range of the mask) with 100 iterations. For the sake of fairness, it is noted that we set the total variation loss in AdvMask and the stealthiness loss to a similar order of magnitude. The experimental results are shown in Table \ref{tab:mul_targets}, from which we can find that our SASMask can deceive the FR model with the best ASR compared to the baseline attacks, reducing the recognition rate from about one hundred percent up to about zero percent with the highest SSIMs. Besides, as for the baseline attacks, randomly perturbed masks and AdvPatch attacked masks (even though with very large perturbation, like $\text{radius}=28$) rarely work towards adversarially attacking the FR model. Some adversarial masked face examples can be seen in Fig. \ref{fig:digital_examples}. 

To demonstrate the stability of the proposed method, we further conducted multi-target attack experiments up to 50 targets. The results are shown in Table \ref{tab:multi-target}, from which we can see as the number of targets increase, ASR and SSIM of our method stabilize at around 90.0\% and 0.9160, respectively. However, the counterparts are ineffective or unstable as the number of targets increases.
\begin{figure}[tbp]
\centering
\includegraphics[width=0.85\linewidth]{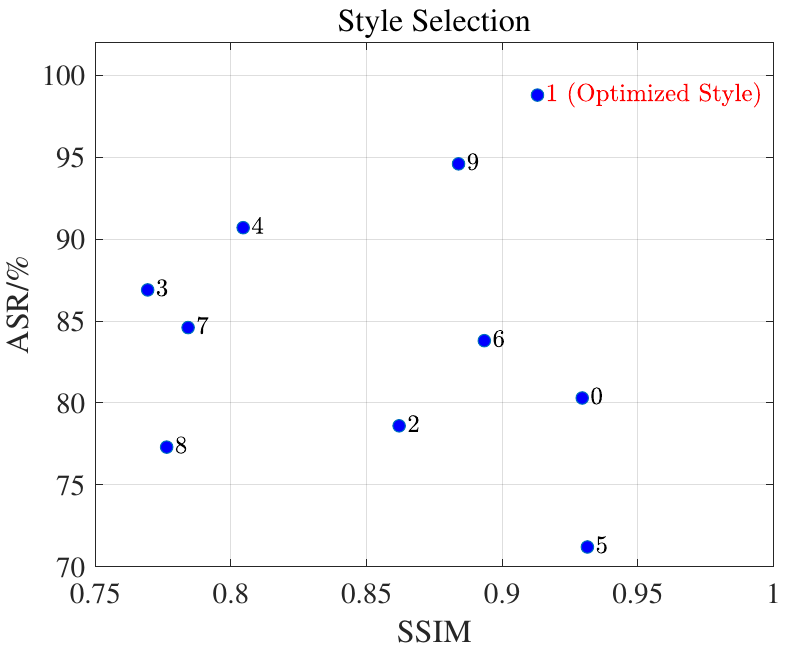}
\caption{ASR and SSIM of adversarial masked face images via manual style selection (style ablation) and style optimization. The optimized style selection is highlighted in red bold font.}
\label{fig:sty_ab}
\vspace{-1em}
\end{figure}
\subsubsection{Black-Box Attack}
Besides, we conducted experiments to evaluate the ASR of our SASMask on FR models that it was not trained on. Specifically, we conducted experiments to examine the attack transferability of different attack methods in the following aspects: model backbone, model head, as well as dataset. Without loss of generality,  Vivica Fox was selected herein as the target identity; PGD-128 and AdvPatch-28 were chosen for these experiments. 

\paragraph{Backbone transferability}
To evaluate the transferable attack performance from one backbone to another, we conducted backbone transferability experiments of different attack methods. Hereof, we fixed the model head as ArcFace and used ResNets with different depths (34, 50, 101), MobileNet, and GhostNet as our model backbones. Besides, for the comparison methods PGD and AdvPatch, we chose the strongest PGD-128 and AdvPatch-28. Note that we only showed the \textbf{average} transfer ASR from one trained backbone to the other four unseen backbones. The experimental results are shown in Table \ref{tab:transfer-backbone}, from which we can find that our attack method has the highest attack transferability between seen and unseen models. Besides, we can see that Random masks, PGD  masks, and AdvPatch masks have little attack transferability from one model backbone to another. For AdvMask and our method, adversarial examples by small backbones have better attack transferability than those by big models, especially for similar backbones (like ResNets). Such observation is consistent with the view presented in \cite{Wu2018}.

\paragraph{Head transferability}
Furthermore, transferable experiments between different model heads were performed to examine the transferability of different attack methods. Hereof, we fixed the model backbone as ResNet-50 and utilized one of the following heads to train the models: ArcFace, CosFace, CircleLoss, CurricularFace, and MagFace. As shown in Table \ref{tab:transfer-head}, our proposed method shows the best transferability between different heads in most cases. Nevertheless, the transferability is not even, e.g., the transferable ASR of ArcFace is much lower than the others, probably because the similarity between ArcFace and the other four heads is lower than that between the other four heads.
\begin{table}[tbp]
\centering
\footnotesize
\caption{ASR without or with different types of masks.}
\label{tab:phy_attack}
\resizebox{\linewidth}{!}{
\begin{tabular}{lccc}
\toprule[1pt]
                    & {Female}  & {Male}    & {All}     \\
\midrule[1pt]
w/o mask             & 2.00\%           & 0.00\%           & 1.00\%           \\
w/ surgical masks    & 4.00\%           & 2.00\%           & 3.00\%           \\
w/ fashionable masks & 6.50\%           & 8.00\%           & 7.25\%           \\
w/ AdvMask           &77.50\%	        &76.00\%	       &76.75\%            \\
\midrule[0.5pt]
\textbf{w/ SASMask} & \textbf{86.00\%} & \textbf{84.50\%} & \textbf{85.25\%} \\
\bottomrule[1pt]
\end{tabular}
}
\vspace{-1em}
\end{table}
\begin{table}[tbp]
\centering
\caption{Scores (score differences from the clean face images) of adversarial face images generated by different methods against online platforms.}
\label{tab:online}
\resizebox{\linewidth}{!}{
\begin{tabular}{lccc}
\toprule[1pt]
          & Baidu         & iFLYTEK       & Tencent        \\
\midrule[0.5pt]
Clean     & 9.39          & 49.26         & 11.01         \\
PGDFace   & 9.41 (0.02\textcolor{red}{$\uparrow$})   & 51.78 (2.52\textcolor{red}{$\uparrow$})  & 10.95 (-0.06\textcolor{green}{$\downarrow$}) \\
PatchFace & 20.24 (10.85\textcolor{red}{$\uparrow$}) & 65.77 (16.51\textcolor{red}{$\uparrow$}) & 21.53 (10.52\textcolor{red}{$\uparrow$}) \\
AdvGlass  & 22.17 (12.78\textcolor{red}{$\uparrow$}) & 67.58 (18.32\textcolor{red}{$\uparrow$}) & 20.95 (9.94\textcolor{red}{$\uparrow$})  \\
AdvHat    & 14.64 (5.25\textcolor{red}{$\uparrow$})  & 59.85 (10.59\textcolor{red}{$\uparrow$}) & 15.19 (4.18\textcolor{red}{$\uparrow$})  \\
AdvMakeup & 28.06 (18.67\textcolor{red}{$\uparrow$}) & 77.24 (27.98\textcolor{red}{$\uparrow$}) & 27.60 (16.59\textcolor{red}{$\uparrow$}) \\
AdvMask   & 31.77 (22.38\textcolor{red}{$\uparrow$}) & \textbf{87.38 (38.12\textcolor{red}{$\uparrow$})} & 35.18 (24.17\textcolor{red}{$\uparrow$}) \\
Ours      & \textbf{33.86 (24.47\textcolor{red}{$\uparrow$})} & 85.86 (36.60\textcolor{red}{$\uparrow$}) & \textbf{38.23 (27.22\textcolor{red}{$\uparrow$})} \\
\bottomrule[1pt]
\end{tabular}
}
\vspace{-1em}
\end{table}
\paragraph{Dataset transferability}
We also evaluated the black-box ASR from one dataset to another dataset. Hereof, we trained different attack methods with the ResNet-50 and ArcFace on the LFW dataset and tested it on the VGGFace2, AgeDB, and CFP datasets. The experimental results are displayed in Table \ref{tab:dataset_trans}, where results on the LFW dataset are the white-box attack results, while the others means the transferable attack results. We can see that even though tested on data of different distributions from the trained ones, our SASMask and AdvMask both show excellent attack transferability, but only our method achieves great stealthiness (high SSIM).

\paragraph{Comparison with other face attack methods}
As mentioned before, we compared our method with five other face attack methods. Here, we used the average transfer ASR of MobileNet. The results are shown in Table \ref{tab:face_attack_comparison}. Comparing with other attacks, our method achieves the best ASR and outweighs the competitors by a large margin. Note that the PGD in PGDFace acts on the whole facial images, for fair comparison and enough definition of images, we set the $\epsilon$ much less than that on the mask region, which is $8/255$.

\subsection{Ablation study}
In this subsection, we performed a series of experiments to analyze the following two aspects of our SASMask: 1) style selection; 2) stealthy loss (i.e., $\mathcal{L}_1$ loss, TV loss, content loss, and style loss).
\begin{figure}[tbp]
\centering
\includegraphics[width=0.99\linewidth]{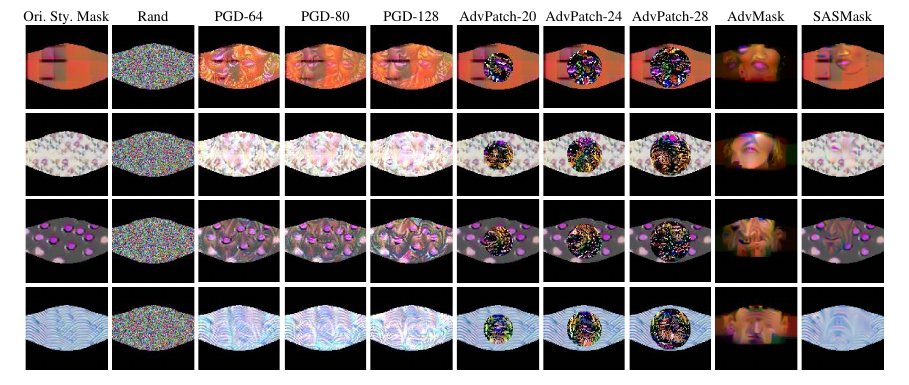}
\caption{Adversarial masks generated by different attack methods. From top to bottom, the target identities are Vivica Fox, Patricia Hearst, Aaron Eckhart and Steve Park, respectively.}
\label{fig:mask_examples}
\vspace{-1em}
\end{figure}
\paragraph{Style optimization}
In this ablation experiment, we compared the attack strength and the stealthiness of the optimized style and the manually selected single style. The results are exhibited in Fig. \ref{fig:sty_ab}, from which we can find that the style optimizer helps us select the optimal adversarial style from the style set that poses the biggest threat to FR models.

\paragraph{Loss ablation}
We further ablated the stealthy loss (i.e., $\mathcal{L}_1$, $\mathcal{L}_{tv}$, $\mathcal{L}_c$ and $\mathcal{L}_s$) to analyze the attack performance of the SASMask. Table \ref{tab:loss_ab} illustrates the loss ablation results, from which we can see that ablating any part of the stealthy loss decreases the SSIM by a great margin without little profit of adversarial strength.

\subsection{Physical attack}
\paragraph{Offline physical attack}
To verify the effectiveness of SASMask in the physical world, we recruited a group of five female and five male participants with the approval of the ethics committee. Firstly, we printed the generated masks on paper and selected six types of masks (one blue surgical mask, four fashionable masks and the adversarial mask) as our counterparts. For each physical experiment, we shot a short video and chose 10 frames as our test data. Thus, we collected 1400 images (700 female images and 700 male images): 100 without any mask, 100 with the surgical mask, 400 with fashionable masks, 400 with AdvMask, and 400 with SASMask. For a type of masked face, the test data of a participant contain 3 left-side images, 3 right-side images and 4 front-side images (10 images in total). Then, we used the MTCNNs \cite{Zhang2016} to detect and align faces. Later, we fed the processed data (some examples are shown in Figs. \ref{fig:mask_examples}, \ref{fig:female_phy} and \ref{fig:male_phy}) into the victim FR model. Here, we used ResNet-50 with ArcFace as the victim FR model. The results are shown in Table \ref{tab:phy_attack}, from which we can see that common masks are hardly aggressive, while AdvMask and our SASMask can deceive the FR models, where our method achieves a higher ASR both for females and males. 

\paragraph{Online platform attack}
Besides, we used our method and the competitors to generate adversarial face images with the LFW test set, and attacked three online commercial FR platforms: Baidu \cite{Baidu2023}, iFLYTEK \cite{iFLYTEK2023}, Tencent \cite{Tencent2023}.  The results are displayed in Table \ref{tab:online}, from which we can see that our method almost achieves the best attack performance among all the listed methods.

%% file: tex/5_conclusion.tex
\begin{figure*}[tbp]
\centering
\includegraphics[width=0.99\linewidth]{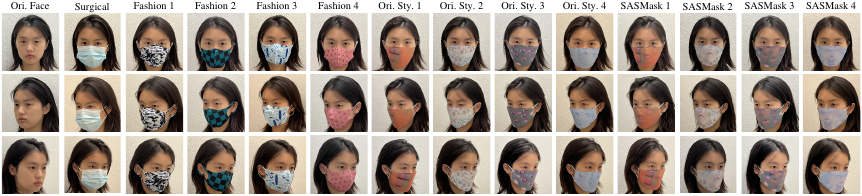}
\caption{Physical (adversarial) masked facial examples of one female participant, where photos in each row are taken from different sides: front, left and right.}
\label{fig:female_phy}
\vspace{-1em}
\end{figure*}
\begin{figure*}[tbp]
\centering
\includegraphics[width=0.99\linewidth]{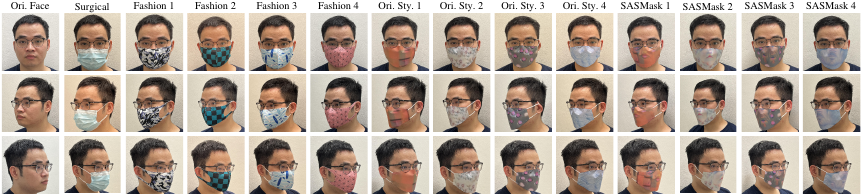}
\caption{Physical (adversarial) masked facial examples of one male participant, where photos in each row are taken from different sides: front, left and right.}
\label{fig:male_phy}
\vspace{-1em}
\end{figure*}
In this paper, we proposed a new stealthy adversarial style mask to attack FR models in both digital and physical settings. Specifically, we trained an adversarial style mask generator that hides adversarial perturbations inside the style masks. Moreover, we proposed to optimize the selection of style to mitigate the sub-optimization of one single style. Compared with existing adversarial works, our method provides both better stealthiness and adversarial strength. Extensive digital experiments, both in white-box settings as well as black-box settings, demonstrated the effectiveness of our proposed SASMask. Furthermore, we printed the SASMask and let participants wear them to attack real-world FR models. In a nutshell, this paper emphasized the vulnerability of deep FR models that they will fail when one wears an elaborated stealthy adversarial style facial mask. 

%% file: tex/7_bio.tex
\begin{IEEEbiography}[{\includegraphics[width=1in,height=1.25in,clip,keepaspectratio]{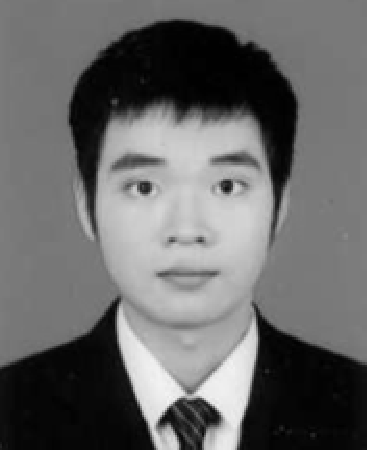}}]{Huihui Gong} received the B.E. degree in Science of Intelligence with the School of Information Science and Technology, Sun Yat-sen University, Guangzhou, China, in 2018. He is currently pursuing the Ph.D. degree in Computer Science at the School of Computer Science, Faculty of Engineering, The University of Sydney, Sydney, Australia. His current research interests include adversarial machine learning and vulnerability detection on software.
\end{IEEEbiography}

\begin{IEEEbiography}[{\includegraphics[width=1in,height=1.25in,clip,keepaspectratio]{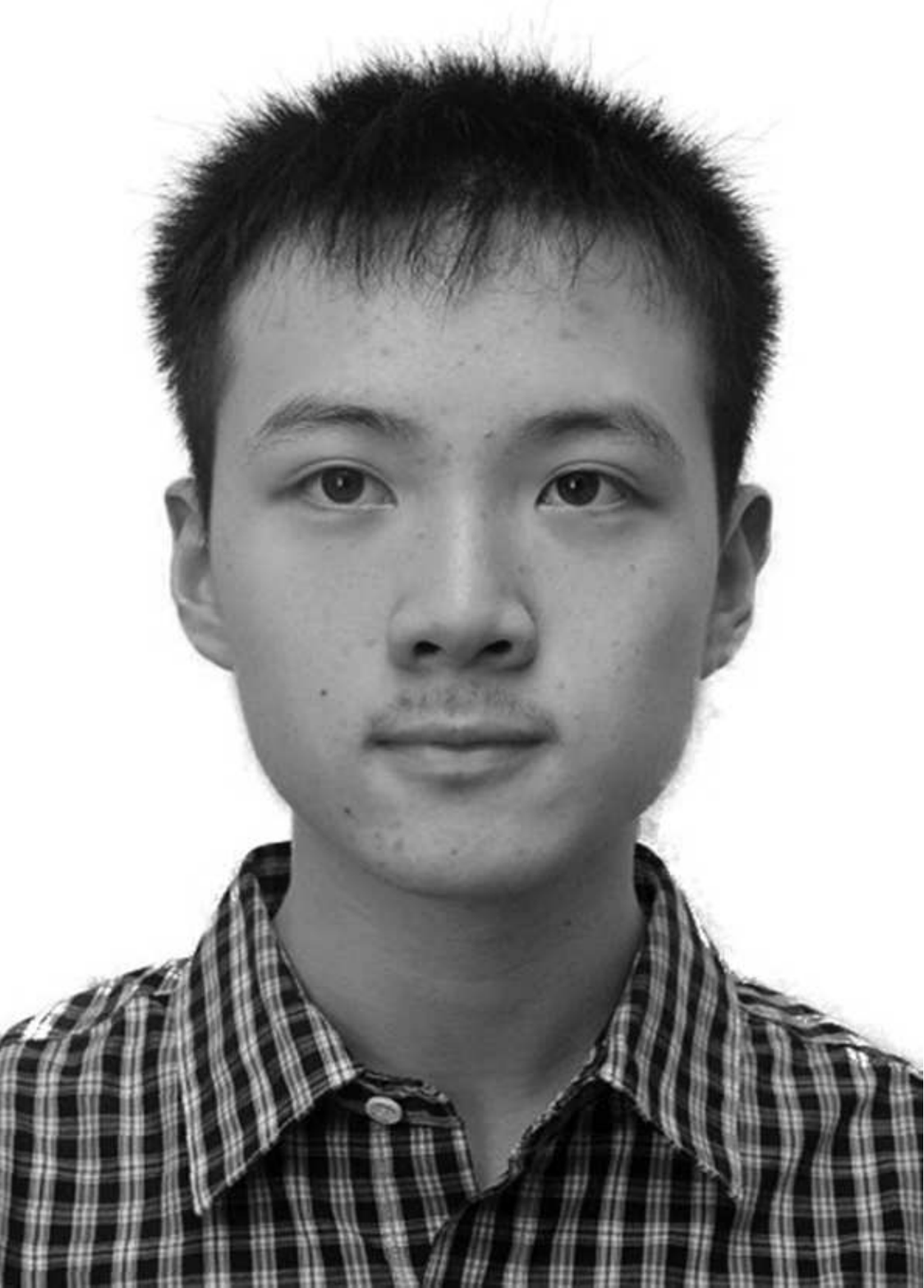}}]{Minjing Dong} is a Phd student at the School of Computer Science, University of Sydney. He received the Mphil degree from University of Sydney. His research interests lie in human behavior analysis, adversarial robustness and neural architecture search.
\end{IEEEbiography}

\begin{IEEEbiography}[{\includegraphics[width=1in,height=1.25in,clip,keepaspectratio]{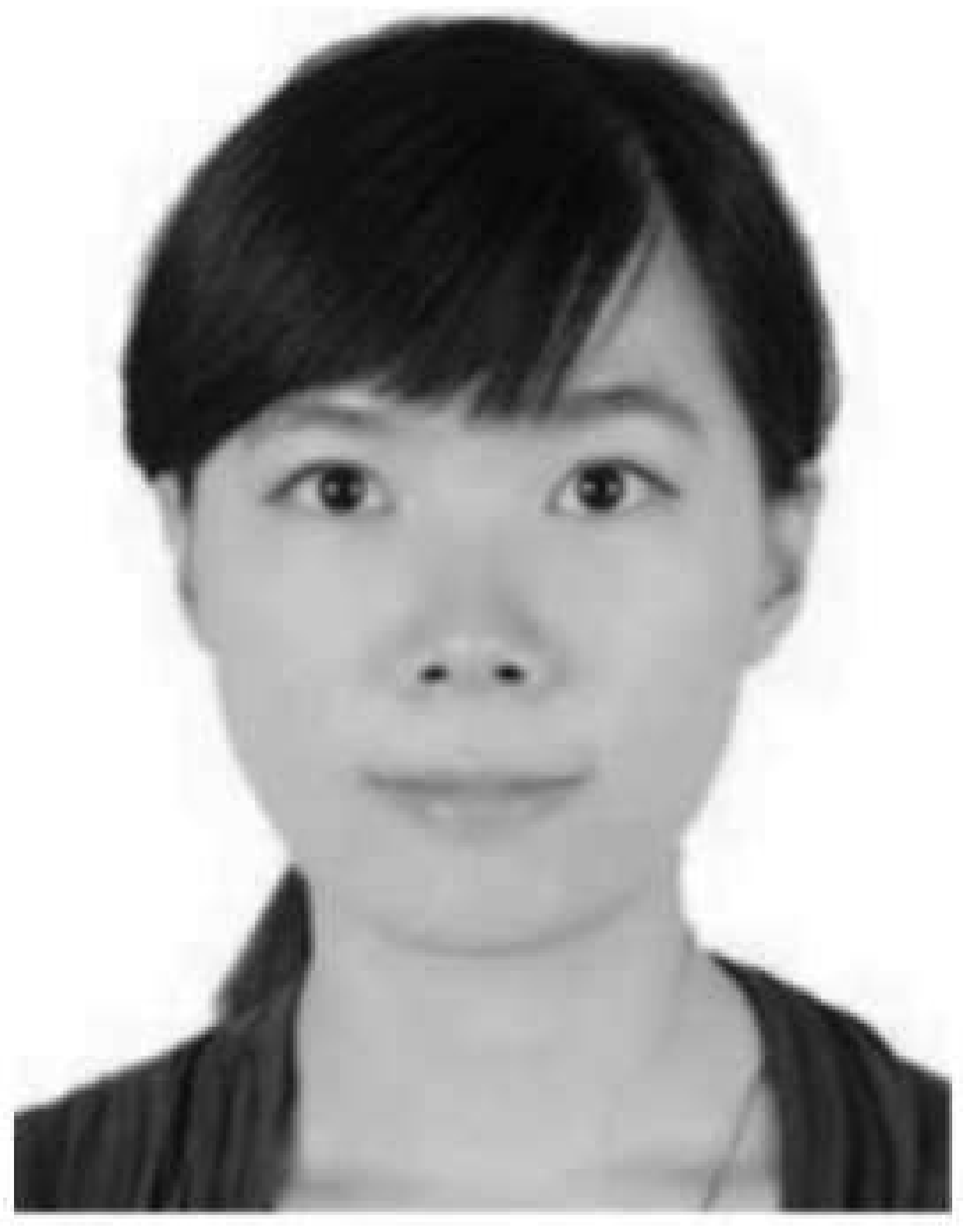}}]{Siqi Ma} received the B.S. degree in computer science from Xidian University, Xi’an, China in 2013 and Ph.D. degree in information system from Singapore Management University in 2018, respectively. She was a research fellow of distinguished system security group from CSIRO and then was a lecturer at University of Queensland. She is currently a senior lecturer of the University of New South Wales, Canberra Campus, Australia. Her research interests include data security, IoT security and software security.
\end{IEEEbiography}

\begin{IEEEbiography}[{\includegraphics[width=1in,height=1.25in,clip,keepaspectratio]{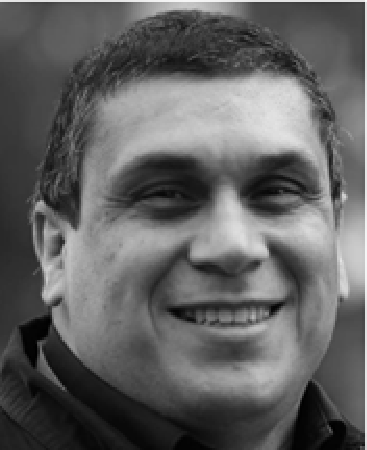}}]{Seyit Camtepe} (Senior Member, IEEE) received the Ph.D. degree from Rensselaer Polytechnic Institute in 2007. He is currently a Principal Research Scientist and the Team Leader with CSIRO Data61. He was with Technische Universitaet Berlin as a Senior Researcher and QUT as a Lecturer. He was among the first to investigate the security of android smartphones and inform society for the rising malware threat. His research interests include autonomous security, malware detection and prevention, smartphone security, applied and malicious cryptography, and CII security.
\end{IEEEbiography}

\begin{IEEEbiography}[{\includegraphics[width=1in,height=1.25in,clip,keepaspectratio]{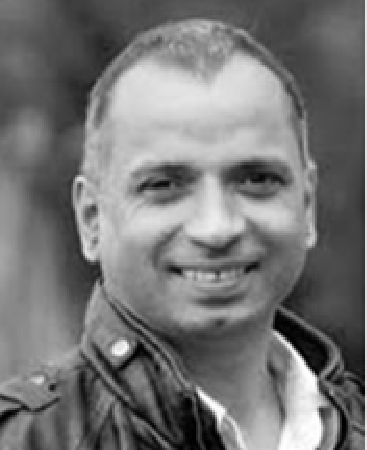}}]{Surya Nepal} is currently a Senior Principal Research Scientist with CSIRO’s Data61. He also leads the distributed systems security group comprising more than 30 research staff and more than 50 postgraduate students. He is also the Team Leader of the Cybersecurity Cooperative Research Centre (CRC), a national initiative in Australia. His main research focus is in the development and implementation of technologies in the area of cybersecurity and privacy, and AI and cybersecurity. He has more than 300 peer-reviewed publications to his credit. He is a member of the Editorial Board of \emph{IEEE Transactions on Services Computing}, \emph{ACM Transactions on Internet Technology}, \emph{IEEE Transactions on Dependable and Secure Computing}, and \emph{Frontiers of Big Data Security Privacy, and Trust}.
\end{IEEEbiography}

\begin{IEEEbiography}[{\includegraphics[width=1in,height=1.25in,clip,keepaspectratio]{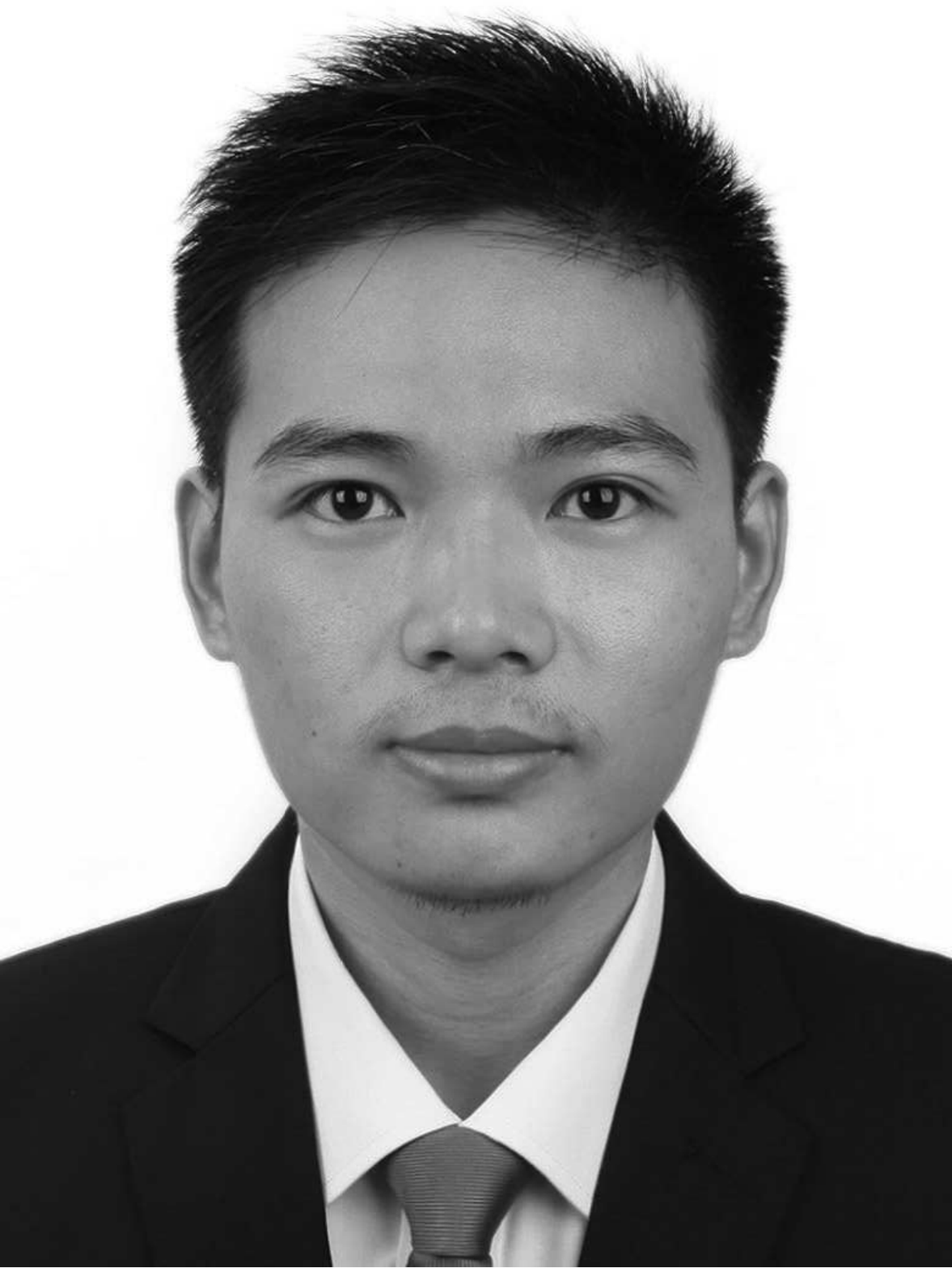}}]{Chang Xu}
 is Senior Lecturer at the School of Computer Science, University of Sydney. He received the University of Sydney Vice-Chancellor’s Award for Outstanding Early Career Research. His research interests lie in machine learning algorithms and related applications in computer vision. He has published over 100 papers in prestigious journals and top tier conferences. He has received several paper awards, including Distinguished Paper Award in AAAI 2023, Best Student Paper Award in ICDM 2022, Best Paper Candidate in CVPR 2021, and Distinguished Paper Award in IJCAI 2018. He served as an area chair of NeurIPS, ICML, ICLR, KDD, CVPR, and MM, as well as a Senior PC member of AAAI and IJCAI. In addition, he served as an associate editor at IEEE T-PAMI, IEEE T-MM, and T-MLR. He has been named a Top Ten Distinguished Senior PC Member in IJCAI 2017 and an Outstanding Associate Editor at IEEE T-MM in 2022.
\end{IEEEbiography}